\documentclass[
]{ceurart}

\usepackage{listings}
\usepackage{float}
\usepackage{placeins}

\begin{document}

\copyrightyear{2026}
\copyrightclause{Copyright for this paper by its authors.
  Use permitted under Creative Commons License Attribution 4.0
  International (CC BY 4.0).}

\conference{}

\title{Specialty-Specific Medical Language Model for Immune-Mediated
Diseases}

\author[1]{Veysel Kocaman}[email=veysel@johnsnowlabs.com]
\author[1]{Gursev Pirge}[email=gursev@johnsnowlabs.com]
\author[1]{Yigit Gul}[email=yigit@johnsnowlabs.com]
\author[1]{Ace Vo}[email=ace@johnsnowlabs.com]
\author[1]{Zhenya Nargizyan}[email=zhenya@johnsnowlabs.com]
\author[1]{David Talby}[email=david@johnsnowlabs.com]

\address[1]{John Snow Labs Inc., Delaware, USA}



\maketitle

\section{Introduction}

Clinical narratives contain details about patient symptoms, diagnostic
reasoning, and therapeutic choices that are often absent from structured
records. Extracting these elements reliably is essential for research
and for improving how clinicians understand and manage disease. Yet
immune-mediated disorders and infections are commonly described in
dense, idiosyncratic free text, making automated interpretation
difficult. Developing methods that can identify and structure clinically
relevant information from these sources remains a central challenge in
healthcare NLP.

Immune-mediated and infectious diseases account for a substantial global
health burden and often present with overlapping manifestations.
Conditions such as rheumatoid arthritis, systemic lupus erythematosus,
multiple sclerosis, and persistent infections evolve over time and
require careful review of longitudinal narratives to distinguish between
immune-driven activity and infectious processes. Symptoms such as
inflammation, fever, or organ dysfunction occur in many of these
disorders and complicate diagnosis. NLP and other AI approaches provide
a way to systematically organize large volumes of unstructured text,
enabling tasks such as early detection, cohort construction, and
assessment of disease trajectories.

Although recent progress has improved clinical NLP capabilities,
general-purpose systems still struggle with the specialized terminology
used in immunology and infectious disease. Variability in how conditions
are described, frequent ambiguity in clinical wording, and the scarcity
of high-quality annotated datasets all reduce the accuracy of existing
models. To address these gaps, we assembled a corpus of 371 case reports
drawn from PubMed, Europe PMC, ScienceDirect, Google Scholar, and
MedRxiv. Two clinical experts annotated the dataset using the Generative
AI Lab~\cite{genai-lab}, defining twelve entity types that capture immune-mediated and
infectious diseases as well as related symptoms and general medical
concepts. Additional paraphrased examples were created to broaden
linguistic diversity and improve the model's robustness.

Using this dataset, we trained several architectures and found that a
Bidirectional Long Short-Term Memory--Convolutional Neural Network with
character-level representations (BiLSTM-CNN-Char) yielded the strongest
performance, reaching an F1 score of 0.89. The work illustrates the
importance of combining domain expertise with tailored NLP methods and
offers a practical framework for extracting immunology- and
infection-related information from free-text documents. Such tools can
support a wide range of downstream applications, including clinical
decision support, disease surveillance, and translational research.

\subsection{Motivation}

To illustrate how the pipeline operates in practice, we applied the
final NER model to a clinical narrative describing a patient with
Systemic Lupus Erythematosus (SLE). SLE provides a useful test case
because of its heterogeneous presentation and the variety of treatments
and laboratory markers commonly documented in clinical notes. The
selected text was processed through the full pipeline, beginning with
entity extraction.

\begin{figure}[htbp]
  \centering
  \includegraphics[width=0.6\linewidth]{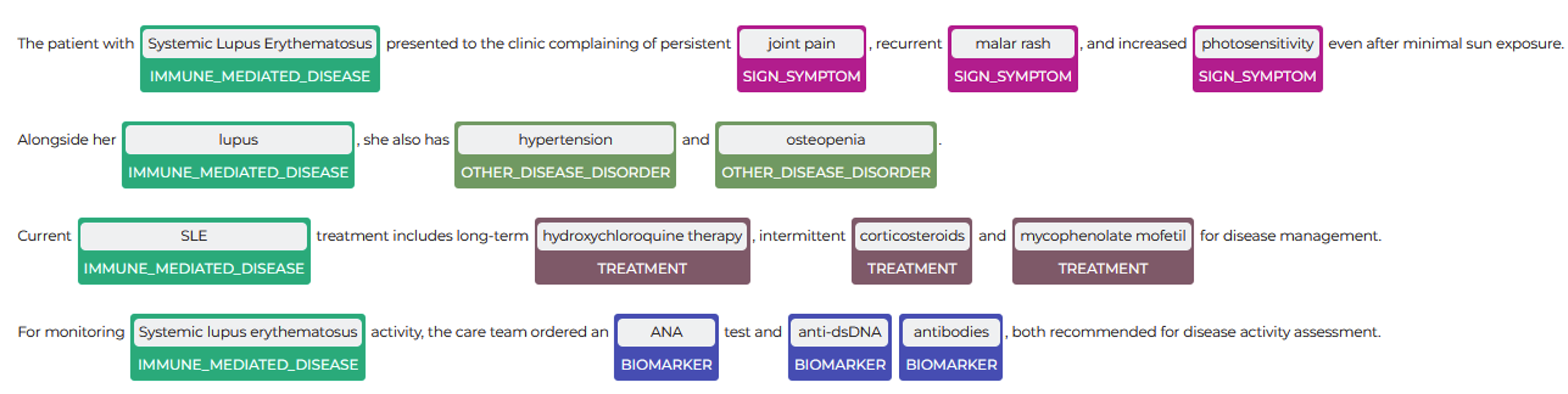}
  \caption{Example of entity recognition performed by the IMD NER model
on a clinical note describing Systemic Lupus Erythematosus (SLE).}
  \label{fig:ner-example}
\end{figure}

Figure~\ref{fig:ner-example} (Appendix) shows the model's output at the NER stage. The system
accurately identified entities corresponding to
Immune\_Mediated\_Disease, Symptom, Treatment, Other\_Disease\_Disorder,
and Biomarker. Key mentions---including Systemic Lupus Erythematosus,
joint pain, malar rash, mycophenolate mofetil, ANA, and anti-dsDNA
antibodies---were detected with correct boundaries and labels. These
results demonstrate that the model handles both common and nuanced
clinical terminology and can reliably convert unstructured text into
structured representations.

\begin{figure}[htbp]
  \centering
  \includegraphics[width=0.6\linewidth]{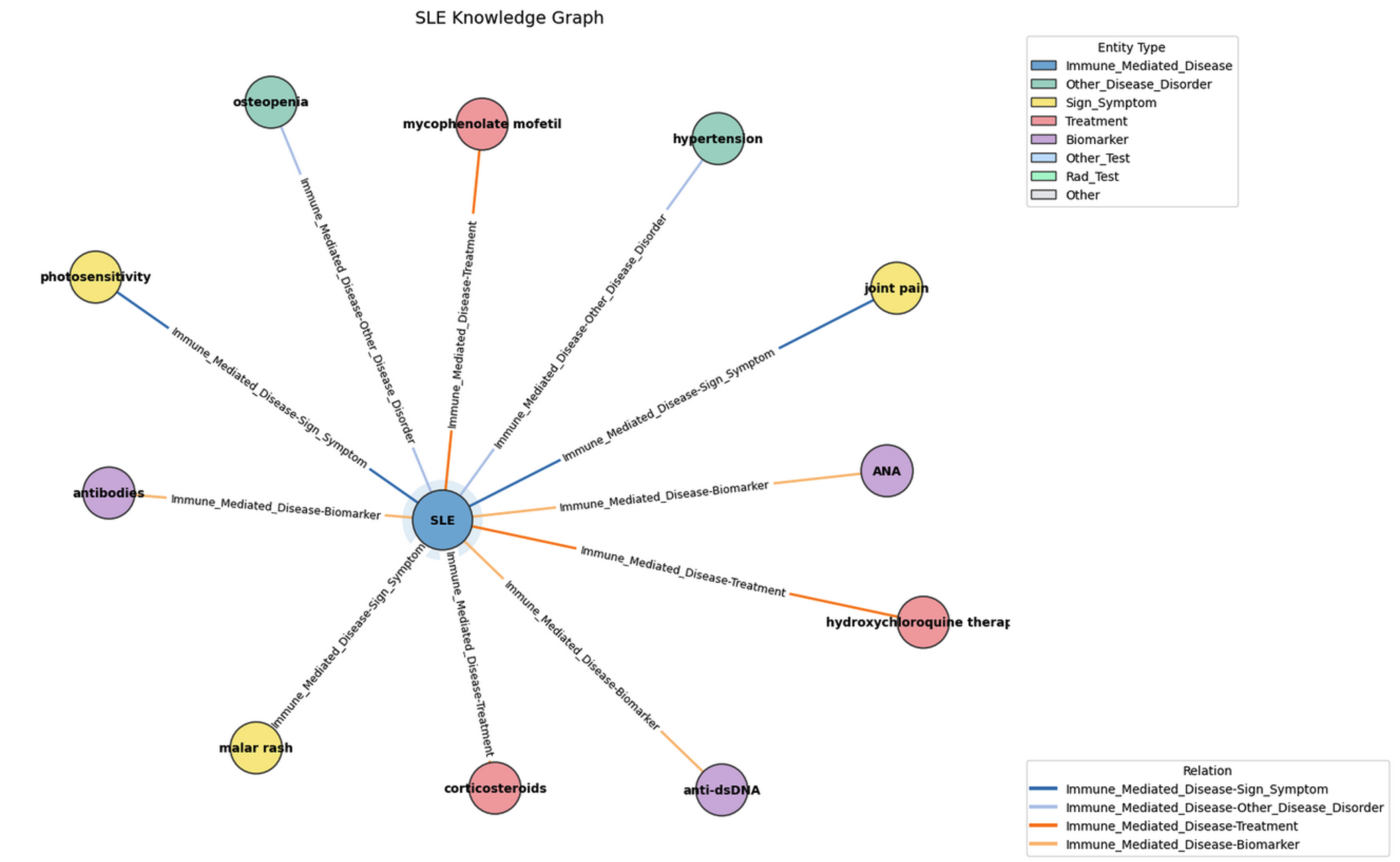}
  \caption{Automatically generated SLE knowledge graph showing
relationships between Immune\_Mediated\_Disease and associated entities.}
  \label{fig:sle-kg}
\end{figure}

In the next step, we applied a Spark NLP Relation Extraction model to
determine how the extracted entities relate to each other. The resulting
SLE Knowledge Graph, generated with NetworkX~\cite{networkx} and shown in Figure~\ref{fig:sle-kg} (Appendix),
maps clinically relevant associations such as
Immune\_Mediated\_Disease--Symptom,
Immune\_Mediated\_Disease--Treatment, and
Immune\_Mediated\_Disease--Biomarker. The graph correctly links SLE to
characteristic manifestations (e.g., photosensitivity, joint pain),
comorbid conditions (hypertension, osteopenia), commonly used therapies
(hydroxychloroquine, corticosteroids), and immunologic markers (ANA,
anti-dsDNA).

Taken together, this example demonstrates how the combined NER and
relation extraction pipeline can automatically transform narrative
clinical descriptions into structured, interpretable knowledge
representations. These outputs can support a range of downstream
applications, including decision-support tools, population of disease
registries, and construction of biomedical knowledge graphs for
immune-mediated disorders. The case study highlights the practical
readiness of the model for real-world clinical and research workflows.

\subsection{Prior Work}

Research at the intersection of infectious and Immune-Mediated Diseases
(IMDs) and clinical NLP has expanded considerably in recent years. Many
studies show that tailoring NER systems to biomedical subdomains can
meaningfully improve tasks such as case identification, cohort
construction, and disease surveillance~\cite{navarro2023}. Disease-focused investigations illustrate how NLP tools can capture
complex IMD phenotypes from free-text sources. Remaki et al.~\cite{remaki2025} designed a multi-component NLP pipeline for immune-mediated
inflammatory diseases (IMIDs) using EHR data, while Kocaman et al.~\cite{kocaman2025} evaluated pretrained clinical NER models across more than 138,000 clinical notes, reporting high precision values. Architectural and linguistic adaptations have also enhanced biomedical NER in languages and subdomains with limited resources~\cite{murakami2023,sun2025,cao2024}.

At a broader level, transformer-based models consistently
outperform rule-driven and statistical systems across extraction,
classification, and phenotyping tasks~\cite{li2022}. However, relatively few efforts have focused
specifically on the combined landscape of infectious and immune-mediated
diseases. This study addresses that gap by presenting a
clinician-annotated corpus and a tailored NER model designed to improve
entity recognition in immunology and infectious disease contexts.

\section{Methods}

The development of the domain-specific NER model followed a sequential
workflow that combined data collection, expert annotation, model
training, and iterative refinement. Clinical narratives were gathered
from several open-access biomedical sources and selected to capture a
range of linguistic styles and diagnostic contexts. Annotation
guidelines were prepared in collaboration with clinical specialists to
standardize how diseases, symptoms, and related entities should be
labeled. The model was trained using the BiLSTM-CNN-Char architecture
available in Spark NLP for Healthcare, which incorporates pre-trained
clinical embeddings. Performance was assessed with standard evaluation
metrics---Precision, Recall, and F1---along with qualitative error
review. Feedback from each training round informed revisions to both the
corpus and annotation scheme, allowing the system to converge toward
stable and generalizable performance.

\subsection{Data Sources}

Clinical case reports and narrative descriptions were collected from
PubMed, ScienceDirect, MedRxiv, and Google Scholar. These repositories
were selected to cover peer-reviewed and preprint literature relevant to
immune-mediated and infectious diseases. To address the limited
availability of rare disease cases and infrequently described symptoms,
we supplemented the dataset with synthetic examples designed to mimic
the phrasing and narrative structure typically seen in case reports.

The IMD dataset was constructed iteratively in eight batches to ensure
broad coverage of immune-mediated and infectious disease narratives
(Table~\ref{tab:data-batches} (Appendix). The seed corpus comprised 52 PubMed case reports, serving as
the foundation for entity definition and guideline refinement.
Subsequent batches expanded the dataset with both real-world and
synthetic materials, balancing clinical diversity and entity
representation. Synthetic samples were generated using multiple large
language model platforms. Later batches introduced long-form and focused
synthetic documents emphasizing complex or rare entities, such as
Geographical\_Location, Fungal\_Infection, and Bacterial\_Infection. In
total, the dataset comprised several hundred annotated texts drawn from
heterogeneous sources, ensuring linguistic variability, topic diversity,
and entity richness suitable for domain-specific NER model training.

\subsubsection{Iterative Improvement Workflow}

\begin{figure}[htbp]
  \centering
  \includegraphics[width=0.4\linewidth]{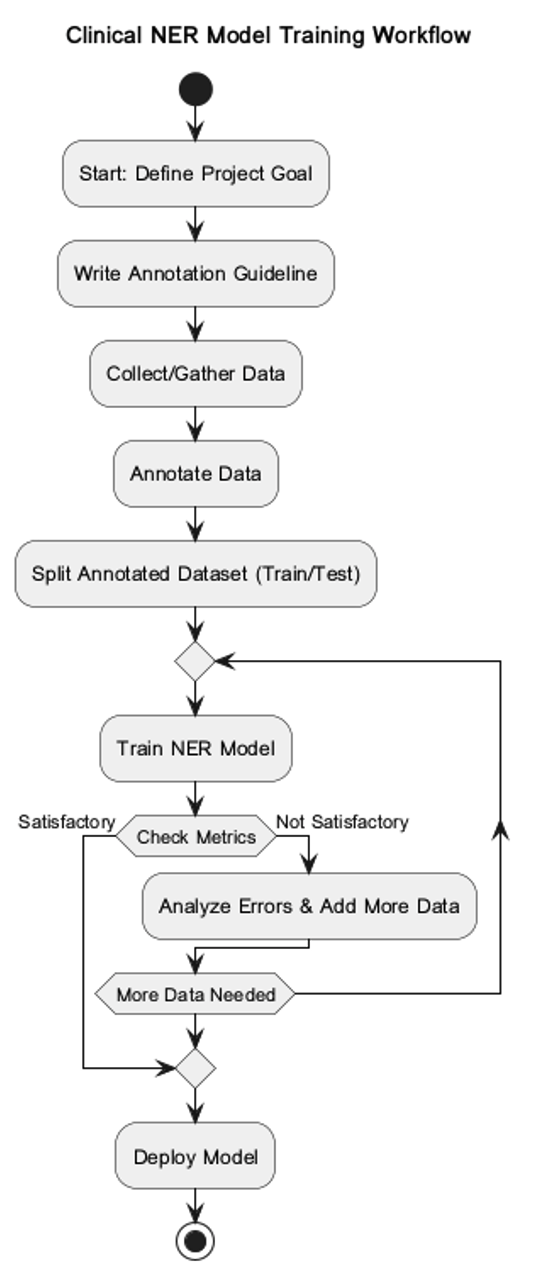}
  \caption{Clinical NER training workflow (full Iteration Loop)}
  \label{fig:workflow}
\end{figure}

Model development relied on an iterative cycle of training, evaluation,
and corpus refinement. After each evaluation round, misclassified
entities and boundary errors were reviewed, and necessary adjustments
were made to the annotation guidelines, especially during the early
stages. Additional examples were added to address recurring error
patterns. The model was then retrained on the expanded dataset. This
loop---summarized in Figure~\ref{fig:workflow} (Appendix)---continued until the performance metrics
plateaued and manual inspection confirmed that the model generalized
well across the diversity of clinical texts.

\subsubsection{Entity Schema}

An Annotation Guideline (AG) was created to define the entity schema for
immune-mediated and infectious disease concepts. The schema consisted of
12 entity types (Table~\ref{tab:entities} (Appendix)), covering diseases, symptoms, treatments,
diagnostic procedures, and related biomedical concepts. The guidelines
specified inclusion and exclusion rules, abbreviation handling, and
boundary conventions. Ambiguous cases were discussed with clinical
specialists, and illustrative examples were provided to help annotators
apply the schema consistently across documents.

\subsubsection{Annotation Process}

Given the importance of high-quality annotations for training NER
models, the corpus was developed through a rigorous expert-driven
process. Two medical doctors conducted the annotations, an approach that
is especially important when working with complex clinical narratives
where subtle distinctions influence label accuracy. Achieving strong
inter-annotator agreement (IAA) required not only clinical expertise but
also clear and well-designed annotation guidelines~\cite{boguslav2017}.

Annotation was performed using the John Snow Labs Generative AI Lab~\cite{genai-lab},
a secure and specialized environment for producing
high-quality healthcare NLP training data. The platform combines
AI-assisted pre-annotation with human review, allowing annotators to
validate and correct suggested labels. For each batch, the model trained
on earlier data provided preliminary labels, which were then reviewed
and refined by the clinical experts. This human-in-the-loop workflow
reduced manual effort while preserving expert-level accuracy. The
platform automatically tracked versioning, annotation consistency, and
IAA across batches, ensuring a transparent and reproducible annotation
process.

\subsubsection{Model Architecture and Training}

The NER model was trained using the \emph{NerModelApproach} annotator~\cite{medicalner-doc}
from Spark NLP. This approach employs a neural architecture
that integrates Character-level Convolutional Neural Networks
(Char-CNNs), Bidirectional Long Short-Term Memory (BiLSTM) layers, and a
Conditional Random Field (CRF) classifier~\cite{kocaman2020,kocaman2022}. The Char-CNN
component captures morphological features and subword patterns, whereas
the BiLSTM layers model contextual dependencies across sequences. The
CRF layer enforces valid tag transitions and improves consistency in the
final label sequence. Hyperparameters were tuned empirically, and the
final configuration is summarized in Table~\ref{tab:hyperparams} (Appendix). The trained model served
as the principal entity extraction component within the pipeline.

\subsubsection{Evaluation Metrics}

Model performance was evaluated using Precision, Recall, and F1-score
computed through the classification\_report function in scikit-learn~\cite{sklearn-doc}.
Metrics were calculated at the entity level to assess label
accuracy and boundary detection. Micro-, macro-, and weighted averages
were reported to provide complementary perspectives on performance
across frequent and infrequent entity types.

Evaluation was performed on a held-out test set comprising 20\% of the
annotated corpus. This subset remained unseen during model development
and hyperparameter tuning. Predictions were compared with the
gold-standard annotations to generate entity-specific and aggregate
performance scores. Error analysis followed each training batch, with
annotators reviewing misclassified spans and incomplete labels.
Corrections were integrated into the corpus to improve annotation
consistency, contributing to performance gains in later training stages.

\section{Results}

\subsection{Dataset Overview}

The final annotated dataset contained 371 clinical and biomedical
narratives covering a broad range of immune-mediated and infectious
disease presentations. These included formal case reports, shorter
clinical descriptions, and synthetic documents created to increase
variability in phrasing and context. Across eight development batches,
the corpus grew to approximately 149,000 tokens with about 22,000
labeled entities distributed across twelve entity categories (Table~\ref{tab:entity-dist} (Appendix)).

\begin{figure}[htbp]
  \centering
  \includegraphics[width=0.6\linewidth]{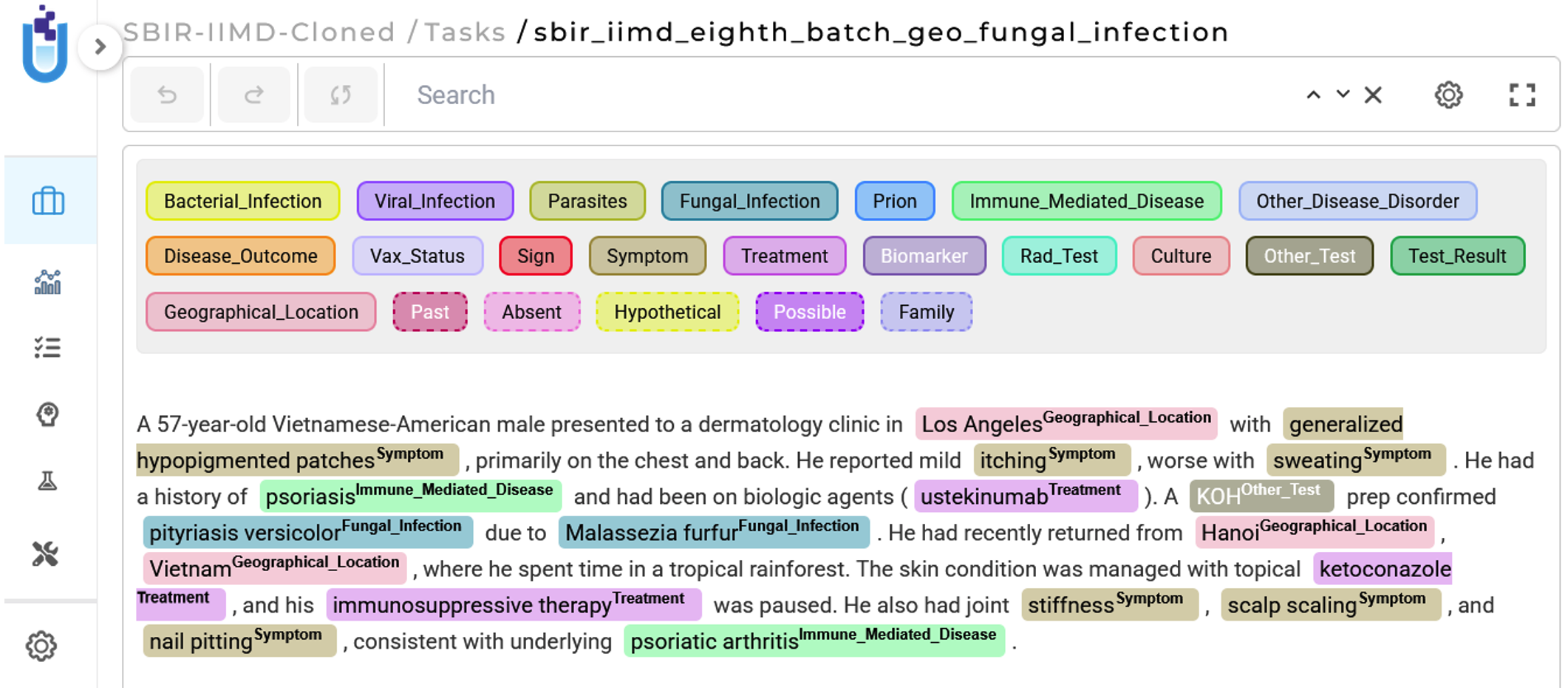}
  \caption{Example annotated case report in Generative AI Lab showing
multiple entity types.}
  \label{fig:annotation-example}
\end{figure}

Figure~\ref{fig:annotation-example} (Appendix) shows an example annotation screen from the Generative AI Lab.
Color-coded labels reflect the project's entity schema and allow
annotators to verify span boundaries directly within the clinical
narrative. The example also illustrates how overlapping or co-occurring
disease processes---common in immune and infectious disorders---were
consistently identified.

\subsection{Annotation Process Metrics}

Table~\ref{tab:annotation-metrics} (Appendix) summarizes statistics captured by the annotation platform,
including task length, annotation time, editing frequency, and
post-processing activity. These metrics provide insight into annotator
workload and the degree of correction required across batches. The
inter-annotator agreement (IAA) score of 89\% indicates strong
consistency between reviewers and reflects the clarity of the guidelines
as well as the benefit of clinical expertise in resolving ambiguous
cases.

\subsection{Model Performance}

The final IMD NER model achieved solid performance across entity types,
with a macro-average F1-score of 0.89 and a micro-average of 0.88 (Table~\ref{tab:performance}).
These results suggest that the model handled both common and less
frequent entity categories reliably and maintained stable boundary
detection across diverse sentence structures and writing styles.

\begin{table}[htbp]
  \caption{IMD NER model accuracy results.}
  \label{tab:performance}
  \begin{tabular}{p{0.35\linewidth}cccc}
    \toprule
    \textbf{Category} & \textbf{Precision} & \textbf{Recall} & \textbf{F1-Score} & \textbf{Support} \\
    \midrule
    Bacterial\_Infection & 0.93 & 0.95 & 0.94 & 439 \\
    Biomarker & 0.80 & 0.91 & 0.85 & 183 \\
    Fungal\_Infection & 0.96 & 0.98 & 0.97 & 109 \\
    Geographical\_Location & 0.96 & 0.97 & 0.96 & 377 \\
    Immune\_Mediated\_Disease & 0.92 & 0.87 & 0.89 & 330 \\
    Other\_Disease\_Disorder & 0.87 & 0.74 & 0.80 & 627 \\
    Other\_Test & 0.85 & 0.88 & 0.86 & 892 \\
    Rad\_Test & 0.82 & 0.96 & 0.89 & 194 \\
    Symptom & 0.86 & 0.89 & 0.87 & 1562 \\
    Test\_Result & 0.89 & 0.87 & 0.88 & 530 \\
    Treatment & 0.85 & 0.94 & 0.89 & 488 \\
    Viral\_Infection & 0.97 & 0.81 & 0.88 & 247 \\
    \midrule
    \textbf{micro avg} & \textbf{0.86} & \textbf{0.90} & \textbf{0.88} & \textbf{5978} \\
    \textbf{macro avg} & \textbf{0.88} & \textbf{0.90} & \textbf{0.89} & \textbf{5978} \\
    \textbf{weighted avg} & \textbf{0.87} & \textbf{0.90} & \textbf{0.88} & \textbf{5978} \\
    \bottomrule
  \end{tabular}
\end{table}

Entity-level analysis showed particularly high performance on
Fungal\_Infection (F1 = 0.97), Bacterial\_Infection (0.94), and
Geographical\_Location (0.96). These categories contain relatively
distinct terminology, which may have contributed to the strong scores.
More variable categories such as Biomarker (0.85) and
Other\_Disease\_Disorder (0.80) showed modestly lower performance but
still achieved useful levels of accuracy. Overall, the entity-level
patterns align with expectations given the heterogeneity in the
underlying clinical texts.

\subsection{Iterative Training Impact}

\begin{figure}[htbp]
  \centering
  \includegraphics[width=0.65\linewidth]{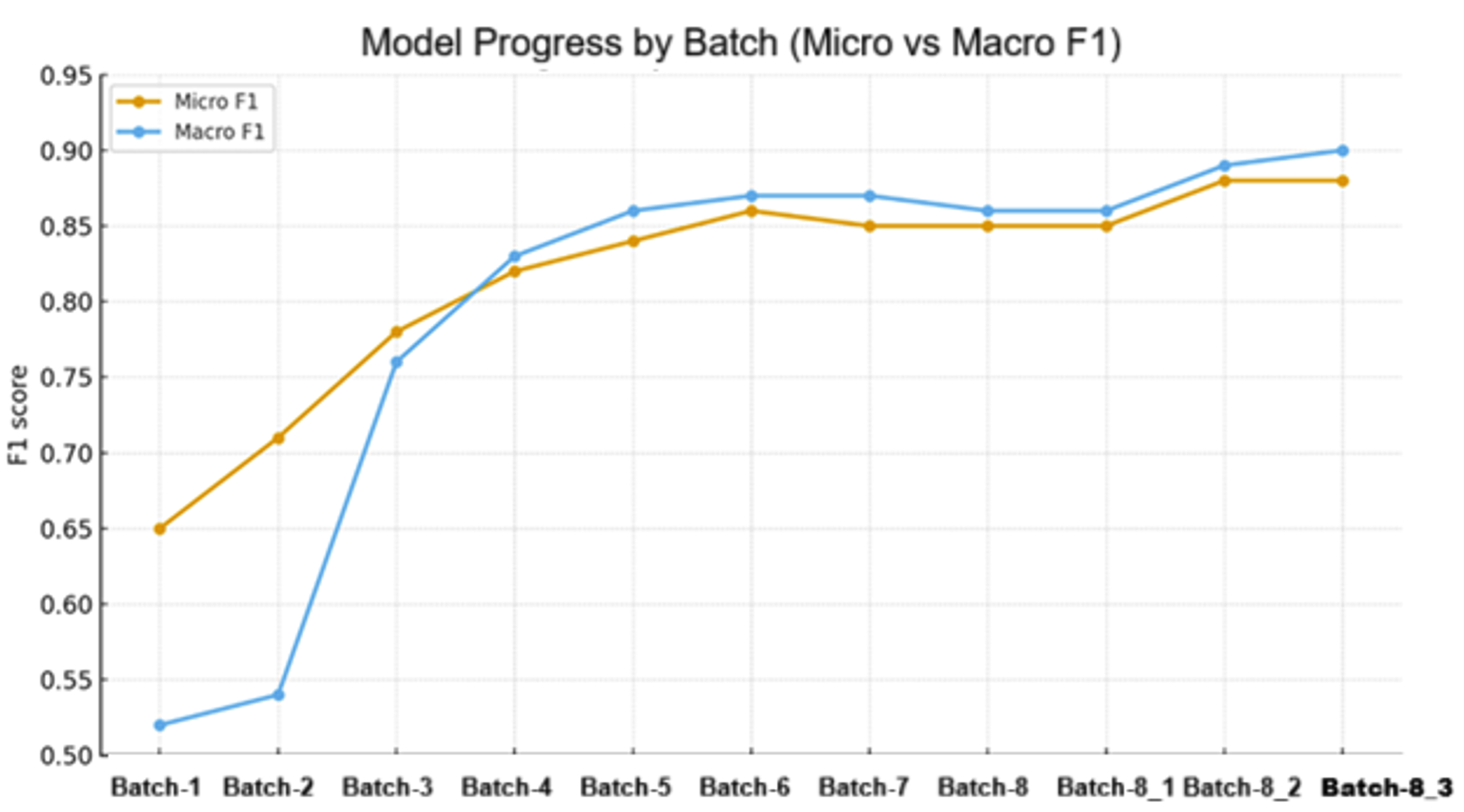}
  \caption{Progress of model performance across annotation batches.}
  \label{fig:training-progress}
\end{figure}

Figure~\ref{fig:training-progress} illustrates the progression of F1-scores across the eight
training batches. Both micro and macro F1 improved steadily as
additional annotated material was incorporated, rising from early values
around 0.65 in Batch 1 to above 0.85 by Batch 7. Macro-F1 increased more
rapidly during the initial rounds, likely due to the addition of
previously underrepresented entity categories. In later stages, the two
curves began to converge, indicating that performance gains had begun to
level off and that the dataset had reached a level of coverage
sufficient for robust generalization.

Overall, the results show that iterative annotation, expert review, and
repeated training cycles substantially strengthened model performance.
The improvement trajectory highlights the practical value of close
collaboration between clinicians and data scientists when developing
specialized clinical NER systems.

\subsection{Comparison of Model Architectures}

To assess how modeling choices influence entity recognition, we trained
several architectures on the IMD corpus and compared them with two
zero-shot systems from the John Snow Labs Healthcare NLP library to
measure the benefit of domain-specific fine-tuning.

Three groups of architectures (shown in Table~\ref{tab:architectures} (Appendix)) were evaluated:
MedicalNER models (BiLSTM--CNN--Char) with BERT~\cite{alsentzer2019} and BioBERT~\cite{lee2020} embeddings, a transformer-based model (BertForTokenClassification~\cite{bert-token-class}) fine-tuned on the IMD dataset, and zero-shot NER models~\cite{zeroshot-ner-doc} applied without corpus-specific adaptation.

Performance for all models is reported in Table~\ref{tab:model-comparison}, which summarizes
precision, recall, and F1-scores under the same evaluation protocol.
This comparison provides a direct view of how traditional sequence
models, transformer-based systems, and zero-shot approaches handle the
linguistic and conceptual complexity of immune-mediated and infectious
disease narratives.

\begin{table}[htbp]
  \caption{Performance metrics.}
  \label{tab:model-comparison}
  \begin{tabular}{p{0.5\linewidth}ccc}
    \toprule
    \textbf{Embeddings / Model} & \textbf{Precision} & \textbf{Recall} & \textbf{F1-score} \\
    \midrule
    BERT For Token Classifier & 0.79 & 0.83 & 0.81 \\
    embeddings\_clinical & \textbf{0.88} & \textbf{0.90} & \textbf{0.89} \\
    bert\_l2\_h256\_uncased & 0.82 & 0.82 & 0.82 \\
    bert\_l2\_h512\_uncased & 0.82 & 0.84 & 0.83 \\
    bert\_l6\_h256\_uncased & 0.81 & 0.84 & 0.82 \\
    bert\_l10\_h768\_uncased & 0.80 & 0.87 & 0.83 \\
    biobert\_pubmed\_base\_cased & 0.84 & 0.89 & 0.86 \\
    biobert\_v1.1\_pubmed & 0.85 & 0.87 & 0.86 \\
    biobert\_pubmed\_base\_cased\_v1.2 & 0.84 & 0.89 & 0.86 \\
    biobert\_clinical\_base\_cased & 0.85 & 0.86 & 0.85 \\
    Zero-Shot Generic & 0.36 & 0.36 & 0.36 \\
    Zero-Shot JSL Healthcare & 0.38 & 0.42 & 0.40 \\
    \bottomrule
  \end{tabular}
\end{table}

The results demonstrate clear performance differences across
architectures and embedding strategies. Clinical and biomedical
embeddings improved accuracy compared to standard BERT token
classification, with embeddings\_clinical achieving the best overall
results (F1 = 0.89). BioBERT variants (F1 $\approx$ 0.86) also performed
strongly, while reduced-dimension BERT models (e.g.,
bert\_l2\_h256\_uncased) were competitive at lower computational cost. The
fine-tuned BertForTokenClassification model was reasonable (F1 = 0.81)
but limited by lack of domain adaptation. Zero-shot models
underperformed (F1 $\leq$ 0.40), and even the healthcare-specific variant
struggled with multi-token boundaries and immunology-specific
terminology.

Overall, domain-adapted embeddings combined with supervised fine-tuning
are required for specialized biomedical NER, while zero-shot approaches
remain insufficient for clinically nuanced terms in immunology and
infectious disease contexts.

\subsection{LLM-Assisted Comparative Analysis}

We evaluated a state-of-the-art large language model (GPT‑5.1) on the same
test set to provide a comparative perspective. GPT-5.1 achieved unexpectedly low
performance across entity types, with a macro-average F1-score of 0.59
and a micro-average of 0.58 (Table~\ref{tab:llm-performance}).

\begin{table}[htbp]
  \caption{ChatGPT-5.1 performance metrics.}
  \label{tab:llm-performance}
  \begin{tabular}{p{0.35\linewidth}cccc}
    \toprule
    \textbf{Category} & \textbf{Precision} & \textbf{Recall} & \textbf{F1-Score} & \textbf{Support} \\
    \midrule
    Bacterial\_Infection & 0.58 & 0.62 & 0.60 & 439 \\
    Biomarker & 0.38 & 0.29 & 0.33 & 183 \\
    Fungal\_Infection & 0.63 & 0.76 & 0.69 & 109 \\
    Geographical\_Location & 0.89 & 0.83 & 0.86 & 377 \\
    Immune\_Mediated\_Disease & 0.65 & 0.72 & 0.68 & 330 \\
    Other\_Disease\_Disorder & 0.55 & 0.59 & 0.57 & 627 \\
    Other\_Test & 0.55 & 0.51 & 0.53 & 892 \\
    Rad\_Test & 0.70 & 0.61 & 0.66 & 194 \\
    Symptom & 0.60 & 0.56 & 0.58 & 1562 \\
    Test\_Result & 0.51 & 0.57 & 0.54 & 530 \\
    Treatment & 0.52 & 0.57 & 0.54 & 488 \\
    Viral\_Infection & 0.51 & 0.55 & 0.53 & 247 \\
    \midrule
    \textbf{micro avg} & \textbf{0.58} & \textbf{0.58} & \textbf{0.58} & \textbf{5978} \\
    \textbf{macro avg} & \textbf{0.59} & \textbf{0.60} & \textbf{0.59} & \textbf{5978} \\
    \textbf{weighted avg} & \textbf{0.59} & \textbf{0.58} & \textbf{0.58} & \textbf{5978} \\
    \bottomrule
  \end{tabular}
\end{table}

Although the prompt included detailed entity definitions and extraction
examples, GPT‑5.1 performed substantially worse than the dedicated NER
model, especially for lexically diverse and overlapping categories
(e.g., Biomarker, Other\_Test, Test\_Result, Treatment). This indicates
that prompt engineering alone could not enforce corpus-specific span
consistency, likely due to missing task-specific fine‑tuning and a
mismatch with the target label set.

\section{Discussion}

\subsection{Principal Results}

This work describes the development of a domain-specific NER model
designed to extract clinically relevant information related to
immune-mediated and infectious diseases from narrative medical text.
Several model families were evaluated, including the MedicalNER
architecture with different clinical and BERT-based embeddings, a
fine-tuned BERT Token Classification model, and two zero-shot NER
systems available through the John Snow Labs Healthcare NLP library.

Models trained with healthcare-adapted embeddings---such as
embeddings\_clinical and the BioBERT variants---consistently
outperformed their counterparts. Their advantage reflects the value of
pretraining on domain-dense corpora, which better captures the
terminology and expression patterns characteristic of immunology and
infectious disease. The generic BERT Token Classification model achieved
reasonable performance but struggled with more complex or highly
specialized entities. The zero-shot systems were able to detect broad
clinical concepts but showed clear limitations when confronted with
immunology-specific terminology or multi-token biomedical entities.

Performance improved steadily across annotation and retraining rounds,
underscoring the role of expert supervision in refining model behavior.
The combination of human review, synthetic data generation, and
domain-specific embeddings ultimately produced a model capable of
accurate and stable entity recognition across a wide range of biomedical
expressions.

As an additional reference baseline, we prompted ChatGPT-5.1 and
evaluated using the same entity definitions and example-guided
instructions. However, the LLM achieved substantially lower performance.
This suggests that, in this setting, prompt-only extraction is not a
substitute for supervised domain-adapted NER.

\subsection{Comparison with Prior Work}

The results are consistent with previous findings in biomedical NLP
showing that models equipped with domain-specific embeddings and
fine-tuned on specialized corpora outperform more general architectures.
Earlier work with BioBERT, ClinicalBERT, and PubMed-trained transformers
similarly demonstrated gains in precision and recall when applied to
clinical or immunology-focused datasets. The superior performance of
healthcare-specific embeddings observed here aligns closely with these
reports.

Although zero-shot NER systems are often proposed as practical solutions
in scenarios with limited annotated data, our results reaffirm their
limitations for tasks requiring fine-grained clinical distinction. Their
weaker performance on immunology-related terminology mirrors
observations in other studies, which have noted that general-purpose
zero-shot NER models lack sufficient exposure to specialized biomedical
vocabulary.

\subsection{Future Work}

Several avenues remain for advancing this line of research. Expanding
the annotated dataset with documents from multiple
healthcare settings or different languages could reduce corpus-specific
bias and improve generalizability. Multi-task learning frameworks that
integrate NER with assertion detection, relation extraction, or document
classification may offer a more holistic approach to clinical
information extraction. Future work will explore hybrid pipelines that
combine LLMs and supervised NER, semi-supervised and few-shot techniques for rare diseases, and evaluation within real-world
workflows such as cohort identification, decision-support tools, or
automated disease registry construction.

\subsection{Conclusions}

This study demonstrates that combining domain-adapted embeddings,
comparative architectural evaluation, and human-in-the-loop refinement
yields a high-performing NER system tailored to immune-mediated and
infectious disease narratives. The consistent improvements observed
across iterative training cycles highlight the effectiveness and
adaptability of the approach. By converting unstructured clinical
narratives into structured biomedical information, the system provides a
scalable foundation for downstream applications in research and
precision healthcare.

\begin{acknowledgments}
This project has been funded in whole or in part with Federal funds from
the National Institute of Allergy and Infectious Diseases, National
Institutes of Health, Department of Health and Human Services, under
Contract №75N93024C00010.
\end{acknowledgments}

\section*{Declaration on Generative AI}
The author(s) did not use any generative AI tools or services in the preparation of this work.

\bibliography{specialty-medlm-imd-refs}

\newpage
\appendix

\section*{Appendix}

\section{Figures}

\newpage

\clearpage
\section{Tables}

\begin{table}[htbp]
  \caption{Overview of data sources and batches for the IMD dataset.}
  \label{tab:data-batches}
  \begin{tabular}{p{0.1\linewidth}p{0.5\linewidth}p{0.1\linewidth}p{0.3\linewidth}}
    \toprule
    \textbf{Batch} & \textbf{Source(s)} & \textbf{Count} & \textbf{Focus / Notes} \\
    \midrule
    1 & PubMed & 52 & Initial corpus of IMD-related cases. \\
    2 & Synthetic (perplexity.ai, DeepSeek) & 20 & Generated to augment entity diversity. \\
    3 & PubMed & 22 & Supplementary disease-specific texts. \\
    4 & PubMed, ScienceDirect, MedRxiv & 32 & Mostly PubMed; added variation in style. \\
    5 & PubMed, ScienceDirect, CDC.org & 30 & Broader coverage of infectious comorbidities. \\
    6 & Synthetic (claude.ai, perplexity.ai) & 58 & Short clinical narratives: IMD entity focus. \\
    7 & PubMed, Msard-journal, ClinicaTerapeutica.it, EuropeanReview.org, Inaactamedica.org, Perplexity.ai & 108 & Emphasis on IMD entities; 10 synthetic cases. \\
    8a & manus.im & 19 & Long-form reports; focusing on: IMD, ODD, Symptom. \\
    8b & ChatGPT, manus.im, perplexity.ai & 25 & Targeted entities: Geographical\_Location and Fungal\_Infection. \\
    8c & ChatGPT, claude.ai, groq.com, perplexity.ai, JSL Medical Chatbot & 18 & Targeted entities: Bacterial\_Infection. \\
    \bottomrule
  \end{tabular}
\end{table}

\begin{table}[htbp]
  \caption{Entity Definitions of the Model.}
  \label{tab:entities}
  \begin{tabular}{p{0.35\linewidth}p{0.65\linewidth}}
    \toprule
    \textbf{Entity} & \textbf{Definition} \\
    \midrule
    Bacterial\_Infection & An infection caused by pathogenic bacteria,
identified by the bacterial organism or the disease it produces. \\
    Biomarker & A measurable biological molecule indicating a normal or
abnormal process, condition, or disease. \\
    Fungal\_Infection & An infection caused by pathogenic fungi, identified
by the fungal organism or the disease resulting from fungal
infection. \\
    Geographical\_Location & A specific place or area mentioned in the text,
such as a country, city, or region. \\
    Immune\_Mediated\_Disease & A condition in which the immune system
abnormally targets the body's own tissues, leading to inflammation and
tissue damage. \\
    Other\_Disease\_Disorder & A disease or disorder not caused by
infectious agents or immune system dysfunction. \\
    Other\_Test & A laboratory diagnostic procedure, excluding culture and
radiological tests, used to detect or monitor diseases and health
conditions. \\
    Rad\_Test & A diagnostic imaging procedure that uses radiation or other
imaging methods to visualize internal body structures. \\
    Symptom & A subjective experience reported by a patient that indicates a
possible disease or condition, not directly observable or measurable by
others. \\
    Test\_Result & The outcome or measurement produced by a laboratory,
surgical, or imaging test, used to aid diagnosis, track disease
progression, or assess treatment effectiveness. \\
    Treatment & A therapeutic intervention or procedure aimed at managing or
curing a disease or condition, including medications, therapies, and
surgeries. \\
    Viral\_Infection & An infection caused by a virus, identified by the
viral organism or the disease resulting from viral activity. \\
    \bottomrule
  \end{tabular}
\end{table}

\begin{table}[htbp]
  \caption{Hyperparameters for training -- optimum parameter (range of values).}
  \label{tab:hyperparams}
  \begin{tabular}{p{0.35\linewidth}p{0.65\linewidth}}
    \toprule
    \textbf{Hyperparameter} & \textbf{Optimal Value (Values Tested)} \\
    \midrule
    Batch Size & 8 (4, 16, 32, 64) \\
    Epoch & 16 (16 - 60) \\
    Learning Rate & 0.001 (0.01, 0.0001) \\
    Dropout Rate & 0.5 (0.3 -- 0.7) \\
    Optimizer & Adam \\
    Embedding Size & 200 \\
    \bottomrule
  \end{tabular}
\end{table}

\begin{table}[htbp]
  \caption{Distribution of entities in the dataset.}
  \label{tab:entity-dist}
  \begin{tabular}{p{0.6\linewidth}p{0.4\linewidth}}
    \toprule
    \textbf{Entity} & \textbf{Count} \\
    \midrule
    Symptom & 4,734 \\
    Other\_Test & 4,098 \\
    Test\_Result & 3,548 \\
    Treatment & 3,337 \\
    Other\_Disease\_Disorder & 3,058 \\
    Bacterial\_Infection & 1,433 \\
    Geographical\_Location & 1,338 \\
    Immune\_Mediated\_Disease & 1,119 \\
    Biomarker & 998 \\
    Viral\_Infection & 993 \\
    Rad\_Test & 933 \\
    Fungal\_Infection & 463 \\
    \bottomrule
  \end{tabular}
\end{table}

\begin{table}[htbp]
  \caption{Annotation process statistics.}
  \label{tab:annotation-metrics}
  \begin{tabular}{p{0.85\linewidth}p{0.15\linewidth}}
    \toprule
    \textbf{Annotation Metric} & \textbf{Value} \\
    \midrule
    Average Task Length (characters) & 2,367 \\
    Average Time Annotators Spent on One Task (seconds) & 1,353 \\
    Average Edit Time Annotators Spent on One Task (seconds) & 225 \\
    Average Number of Edits per Task & 2 \\
    Average Number of Tokens for Submitted/Reviewed Tasks & 86.94 \\
    Total Number of Tokens Across Submitted/Reviewed Tasks & 31,298 \\
    Inter Annotator Agreement (IAA) Between Annotators (\%) & 89 \\
    \bottomrule
  \end{tabular}
\end{table}

\begin{table}[htbp]
  \caption{Architectures and embeddings.}
  \label{tab:architectures}
  \begin{tabular}{p{0.25\linewidth}p{0.35\linewidth}p{0.4\linewidth}}
    \toprule
    \textbf{Annotator} & \textbf{Architecture} & \textbf{Embeddings / Backbone / Model} \\
    \midrule
    \multirow{9}{=}{MedicalNER} & \multirow{9}{=}{BiLSTM-CNN-Char} &
    embeddings\_clinical~\cite{embeddings-clinical} \\
    & & biobert\_pubmed\_base\_cased~\cite{biobert-pubmed} \\
    & & biobert\_v1.1\_pubmed~\cite{biobert-v1.1} \\
    & & biobert\_pubmed\_base\_cased\_v1.2~\cite{biobert-v1.2} \\
    & & biobert\_clinical\_base\_cased~\cite{biobert-clinical} \\
    & & bert\_l2\_h256\_uncased~\cite{bert-l2-h256} \\
    & & bert\_l2\_h512\_uncased~\cite{bert-l2-h512} \\
    & & bert\_l6\_h256\_uncased~\cite{bert-l6-h256} \\
    & & bert\_l10\_h768\_uncased~\cite{bert-l10-h768} \\
    BERT For TokenClassifier & Transformer-based architecture for sequence
labeling tasks & Uses BERT's contextualized token representations \\
    PretrainedZeroShotNER & Transformer-based model & zeroshot\_ner\_generic\_large~\cite{zeroshot-generic} \\
    \bottomrule
  \end{tabular}
\end{table}

\end{document}